%% file: root.tex
\documentclass[letterpaper, 10 pt, conference]{IEEEtran}

\IEEEoverridecommandlockouts

\usepackage[letterpaper,left=.75in,right=.75in,top=.75in,bottom=.75in]{geometry}

\usepackage[pdftex]{graphicx}
\usepackage{amsmath}
\usepackage{amssymb}
\usepackage[english]{babel}
\usepackage[utf8]{inputenc}
\usepackage{csquotes}
\usepackage{environ}
\usepackage[T1]{fontenc}
\usepackage{graphics}
\usepackage{hyperref}
\usepackage{multirow}
\usepackage{siunitx}
\usepackage[font=small]{caption}
\usepackage{subcaption}
\usepackage{tikz, pgfplots}
\usepackage{textcomp}
\usetikzlibrary{plotmarks}

\makeatletter
\newsavebox{\measure@tikzpicture}
\NewEnviron{scaletikzpicturetowidth}[1]{%
  \def\tikz@width{#1}%
  \def\tikzscale{1}\begin{lrbox}{\measure@tikzpicture}%
  \BODY
  \end{lrbox}%
  \pgfmathparse{#1/\wd\measure@tikzpicture}%
  \edef\tikzscale{\pgfmathresult}%
  \BODY
}
\makeatother

\graphicspath{{figures/}}
\hypersetup{
colorlinks=true,
linkcolor=black,
citecolor=black,
filecolor=black,
urlcolor=black,
}

\newcommand{\norm}[1]{\left\lVert#1\right\rVert}
\DeclareMathOperator*{\argmin}{arg\,min}

\title{\vspace{.25in}
Object Detection and Classification in Occupancy Grid Maps using Deep Convolutional Networks
\thanks{J.\,B. Frias thanks University of Vigo for funding his research period at Karlsruhe Institute of Technology (KIT), Germany. \textcopyright 2018 IEEE}
}

\author{\IEEEauthorblockN{Sascha Wirges, Tom Fischer and Christoph Stiller}
\IEEEauthorblockA{Mobile Perception Systems Group\\
FZI Research Center for Information Technology\\
Karlsruhe, Germany\\
Email: \{wirges,fischer,stiller\}@fzi.de}
\and
\IEEEauthorblockN{Jesus Balado Frias}
\IEEEauthorblockA{Applied Geotechnologies Research Group\\
University of Vigo\\
Vigo, Spain\\
Email: jbalado@uvigo.es}
}

\begin{document}

\maketitle
\thispagestyle{empty}
\pagestyle{empty}

\input{sections/00_abstract}
\input{sections/01_introduction}
\input{sections/02_related_work}
\input{sections/03_grid_map_processing}
\input{sections/04_training}
\input{sections/05_evaluation}
\input{sections/06_conclusion}

\bibliographystyle{IEEEtran}
\bibliography{root}

\end{document}

%% file: sections/00_abstract.tex
\begin{abstract}
Detailed environment perception is a crucial component of automated vehicles.
However, to deal with the amount of perceived information, we also require segmentation strategies.
Based on a grid map environment representation, well-suited for sensor fusion, free-space estimation and machine learning, we detect and classify objects using deep convolutional neural networks.
As input for our networks we use a multi-layer grid map efficiently encoding 3D range sensor information.
The inference output consists of a list of rotated bounding boxes with associated semantic classes.
We conduct extensive ablation studies, highlight important design considerations when using grid maps and evaluate our models on the KITTI Bird's Eye View benchmark.
Qualitative and quantitative benchmark results show that we achieve robust detection and state of the art accuracy solely using top-view grid maps from range sensor data.
\end{abstract}

%% file: sections/01_introduction.tex
\section{Introduction}
\label{sec:introduction}

\begin{figure}[!ht]
\centering
\def\svgwidth{\columnwidth}
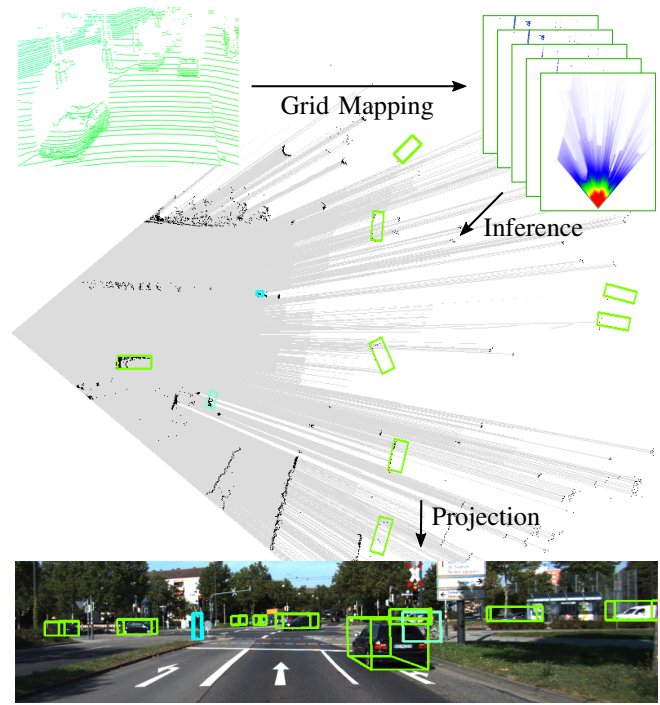
\caption{
System overview.
We transform range sensor measurements to a multi-layer grid map which serves as input for our object detection and classification network.
From these top view grid maps the network infers rotated 3D bounding boxes together with semantic classes.
These boxes can be projected into the camera image for visual validation.
Cars are depicted green, cyclists aquamarin and pedestrians cyan.
}
\label{fig:system_overview}
\end{figure}

We require a detailed environment representation for the safe use of mobile robotic systems, e.g. in automated driving.
To enable higher level scene understanding and decrease computational cost for existing methods, information needs to be further filtered, segmented and categorized.
This task can be accomplished by object detection, shape estimation and classification methods, in the following simply referred to as \textit{object detection}.
Given an input environment representation, the object detector should output a list of oriented shapes and their corresponding most likely semantic classes.

In this work we represent the environment by top-view grid maps, in the following referred to as \textit{grid maps}.
Occupancy grid maps, first introduced in \cite{elfes1989} encode surface point positions and free-space from a point of view in a two-dimensional grid.
As all traffic participants move on a common ground surface one might not require full 3D information but instead represent the scene in 2D with obstacles occupying areas along the drivable path.
Multi-layer grids are well-suited for sensor fusion \cite{nuss2015} and their organized 2D representation enables the use of efficient convolutional operations for deep learning in contrast to sparse point sets.

Compared to the camera image domain there are a few key differences for object detection in grid maps.
While objects in camera images vary in scale due to the projective mapping grid maps represent an orthographic top view composed of metric fixed-size cells.
This representation makes objects scale invariant.
In addition, object projections in camera images might overlap which is not the case for multiple objects in occupancy grid maps.
In this work, we exploit these key differences and adapt modern convolutional object detectors from the camera to the grid map domain.

First, we present an overview on object detection in images and multi-layer grid maps in Section~\ref{sec:related_work}.
After presenting our grid map processing steps in Section~\ref{sec:grid_map_processing} we provide detailed information on the training of our object detectors in Section~\ref{sec:training}.
By making specific design considerations for the grid map domain we are able to train object detectors in an end-to-end fashion and achieve state-of-the-art accuracy at reasonable processing time compared to recent 3D object detection approaches.
These results are discussed in Section~\ref{sec:evaluation} where we show the influence of various parameters and discuss their effects on performance in detail.
Subsequently, we compare the most promising object detection models to recent state-of-the-art approaches on the KITTI bird's eye view benchmark and provide exemplary results for qualitative comparison.
Finally, we conclude our work and propose future plans for object detection in Section~\ref{sec:conclusion}.

%% file: figures/front.pdf_tex
\begingroup%
  \makeatletter%
  \providecommand\color[2][]{%
    \errmessage{(Inkscape) Color is used for the text in Inkscape, but the package 'color.sty' is not loaded}%
    \renewcommand\color[2][]{}%
  }%
  \providecommand\transparent[1]{%
    \errmessage{(Inkscape) Transparency is used (non-zero) for the text in Inkscape, but the package 'transparent.sty' is not loaded}%
    \renewcommand\transparent[1]{}%
  }%
  \providecommand\rotatebox[2]{#2}%
  \ifx\svgwidth\undefined%
    \setlength{\unitlength}{1818.44461596bp}%
    \ifx\svgscale\undefined%
      \relax%
    \else%
      \setlength{\unitlength}{\unitlength * \real{\svgscale}}%
    \fi%
  \else%
    \setlength{\unitlength}{\svgwidth}%
  \fi%
  \global\let\svgwidth\undefined%
  \global\let\svgscale\undefined%
  \makeatother%
  \begin{picture}(1,1.06564127)%
    \put(0,0){\includegraphics[width=\unitlength,page=1]{front.pdf}}%
    \put(0.73331048,0.74220824){\color[rgb]{0,0,0}\makebox(0,0)[lt]{\begin{minipage}{0.16850392\unitlength}\raggedright Inference\end{minipage}}}%
    \put(0.42277096,0.92924944){\color[rgb]{0,0,0}\makebox(0,0)[lt]{\begin{minipage}{0.23952706\unitlength}\raggedright Grid Mapping\end{minipage}}}%
    \put(0.65388079,0.29881511){\color[rgb]{0,0,0}\makebox(0,0)[lt]{\begin{minipage}{0.16850392\unitlength}\raggedright Projection\end{minipage}}}%
    \put(0,0){\includegraphics[width=\unitlength,page=2]{front.pdf}}%
  \end{picture}%
\endgroup%

%% file: sections/02_related_work.tex
\section{Related Work}
\label{sec:related_work}

\subsection{Object Detection Meta-Architectures}
\label{sec:related_work_meta_architectures}
Recently, a notable amount of state-of-the-art object detectors is based on the Faster R-CNN meta-architecture \cite{ren2017}.
In Faster R-CNN detection happens in two stages, a region proposal network (RPN) and a classification and box refinement network.
In the RPN features are extracted from the input and used to predict class-agnostic box candidates in a grid of anchors tiled in space, scale and aspect ratio.
The feature slice corresponding to each box proposal is then sequentially fed into the box classifier.
In the original Faster R-CNN implementation each feature slice is fed into two dense layers before performing classification and box refinement whereas in R-FCN \cite{dai2016} the dense layers are omitted, reducing the amount of computation per region.
In contrast to Faster R-CNN and R-FCN single shot detectors (SSDs) \cite{liu2015} predict bounding boxes and semantic classes with a single feed-forward CNN, significantly reducing inference time but also lowering the overall accuracy.

\subsection{Feature Extractors}
\label{sec:related_work_feature_extractors}
The detection stage input consists of high-level features.
These features may be computed by a deep feature extractor such as Resnet \cite{he2016}, Inception \cite{szegedy2016} or MobileNet \cite{howard2017}.
Resnets implement layers as residual functions, gain accuracy from increased depth and were successfully applied in the ILSVRC and COCO 2015 challenges.
Among other aspects, Inception and MobileNet use factorized convolutions to optimize accuracy and computation time.
With Inception units the depth and width of networks can be increased without increasing computational cost.
MobileNets further reduce the number of parameters by using depth-wise separable convolutions.

\subsection{Object Detection in Aerial Images}
\label{sec:related_work_aerial_images}
Here, we compare the object segmentation task in grid maps to (scale-corrected) satellite or aerial images which has a long research history \cite{hinz2003b, nguyen2007, kluckner2007}.
For example, \cite{nguyen2007} uses 1420 labeled samples in high resolution panchromatic images to train a vehicle detector, reducing false positives by selecting only hypotheses on surfaces semantically classified as streets.
Whereas atmospheric conditions might limit aerial image quality due to camera views far from the scene top view grid maps suffer from occlusions due to a view within the scene.
These problems can either be tackled by fusing multiple measurements from different views or learned environment reconstruction \cite{wirges2018}.
However, \cite{zhao2003} considers the shadows\,/\,occlusions from cars one of the most relevant features (together with the rectangular shape and the windshield layout).

\subsection{KITTI Bird's Eye View Benchmark}
\label{sec:related_work_kitti_bev_benchmark}
Training deep networks requires a comparably large amount of labeled data.
The KITTI Bird's Eye View Evaluation 2017 \cite{geiger2012} consists of 7481 camera images for training and 7518 images for testing as well as corresponding range sensor data represented as point sets.
Training and test data contain 80,256 labeled objects in total which are represented as oriented 3D bounding boxes (7 parameters).
As summarized in Table~\ref{tab:classes}, there are eight semantic classes labeled in the training set although not all classes are used to determine the benchmark result.

\begin{table}[ht]
\centering
\begin{tabular}{c|c|c|c}
\textbf{Class} 			& \textbf{Occurrence} 	& \textbf{Max. length}	& \textbf{Max. width} 	\\
						& \%						& $m$					& $m$					\\
\hline
\textbf{Car} 			& 70.8					& 6.67 					& 2.04 					\\
\textbf{Pedestrian} 		& 11.1					& 1.44 					& 1.20 					\\
\textbf{Van} 			& 7.2					& 6.91 					& 2.52 					\\
\textbf{Cyclist} 		& 4.0					& 2.17 					& 0.93 					\\
\textbf{Truck} 			& 2.7					& 16.79 					& 3.01 					\\
\textbf{Misc} 			& 2.4					& 12.6 					& 2.68					\\
\textbf{Tram} 			& 1.3					& 35.24 					& 2.81 					\\
\textbf{Sitting person} & 0.6					& 1.33 					& 0.78
\end{tabular}
\caption{Semantic classes available in the KITTI Bird's Eye View Evaluation 2017.
Occurrences and max. length\,/\,width are provided for the training set.
In the evaluation vans are not considered car false positives and sitting persons are not considered pedestrian false positives.
}
\label{tab:classes}
\end{table}

Currently, successful benchmark submissions share a two-stage structure comprised of RPN and box refinement and classification network \cite{chen2017, ku2017}.
They first extract features from sensor data, create axis aligned object proposals and perform classification and box regression on the best candidates.
Whereas the region proposals in \cite{chen2017} are based only on grid maps, \cite{ku2017} also incorporates camera images to generate proposals.
To further increase accuracy \cite{ku2017} trains two separate networks for cars and pedestrians/cyclists, respectively.

\subsection{Choice of Input Features}
\label{sec:related_work_input_features}
The choice of grid cell features varies heavily along different publications.
\cite{golovinskiy2009, babahajiani2015} and \cite{chen2017} use the (normalized) number of detections and characteristics derived from detection reflectances.
As the reduction of 3D range sensor information to 2D implies a loss of information features that encode height information might be relevant.
\cite{golovinskiy2009} uses the average height and an estimate of its standard deviation as features while \cite{chen2017} uses four height values, equally sampled in the interval between the lowest and the highest point coordinate of each cell.

There are also higher level features possible.
\cite{hoermann2018} uses evidence measures for occupied and free cells, average velocity and its auto-covariance matrix estimated by a particle filter.
\cite{golovinskiy2009} estimates the standard deviations in the two principle horizontal directions whereas \cite{babahajiani2015} estimates local planarity.
However, as we aim to train object detectors in an end-to-end fashion we do not consider handcrafted features in this work.
On the one hand, it seems sometimes arbitrary to us how certain features are picked and there is no evidence of gaining accuracy when using higher-level features in combination with the training of deep networks.
On the other hand, higher-level features such as velocity estimates might not be available at all times.

\subsection{Box Encoding}
\label{sec:related_work_box_encoding}
Similar to the feature encoding of grid cells there are a variety of different box encodings used in related work.
\cite{chen2017} uses eight 3D points (24 parameters) for box regression and recover the box orientation in direction of the longer box side.
In contrast, \cite{ku2017} uses four ground points and the height of the upper and lower box face, respectively (14 parameters).
They explicitly regress the sine and cosine of orientation to handle angle wrapping and increase regression robustness.
One encoding that needs the minimum amount of 2D box parameters (5) is presented in \cite{jiang2017}.
They represent boxes by two points and one height parameter (5 parameters).

%% file: sections/03_grid_map_processing.tex
\section{Grid Map Processing}
\label{sec:grid_map_processing}

We perform minimal preprocessing in order to obtain occupancy grid maps.
As there are labeled objects only in the camera image we remove all points that are not in the camera's field of view (see Figure~\ref{fig:features}).
We then apply optional ground surface segmentation described in Section~\ref{sec:grid_map_processing_ground_surface_segmentation} and estimate different grid cell features summarized in Section~\ref{sec:grid_map_processing_grid_cell_features}.
The resulting multi-layer grid maps are of size 60\,m\,$\times$\,60\,m and a cell size of either 10\,cm or 15\,cm.

\subsection{Ground Surface Segmentation}
\label{sec:grid_map_processing_ground_surface_segmentation}
Recent approaches create top view images including all available range sensor points.
However, it remains unclear if ground surface points significantly influence the object detection accuracy.
Therefore, we optionally split ground from non-ground points.
As we observed the ground to be flat in most of the scenarios we fit a ground plane to the representing point set.
However, any other method for ground surface estimation can be used as well.
For each scan, we perform nonlinear Least-Squares optimization \cite{ceres} to find the optimal plane parameters
\begin{equation*}
\mathbf{pl}^* = \argmin_{\mathbf{pl}} \sum_{\mathbf{p} \in \mathcal{P}} \rho\left( \norm{\mathbf{e} \left( \mathbf{pl}, \mathbf{p} \right)}^2 \right)
\end{equation*}
which minimize the accumulated point-to-plane error for all points $\mathbf{p}$ of the point set where $\mathbf{e} \left( \mathbf{pl}, \mathbf{p} \right)$ denotes the distance vector between $\mathbf{p}$ and its plane projection point.
The loss function $\rho$ is chosen to be the Cauchy loss with a small scale (5 cm) to strictly robustify against outliers.
We then remove all points from the point set with signed distance below $0.2 \mkern\thinmuskip$m to the plane.

\subsection{Grid Cell Features}
\label{sec:grid_map_processing_grid_cell_features}
We use the full point set or a non-ground subset to construct a multi-layer grid map containing different features.
Inspired by other contributions (e.g. \cite{chen2017, ku2017}) we investigate if there is evidence for better convergence or accuracy by normalizing the number of detections per cell.

\begin{figure}[!ht]
\centering
\begin{subfigure}{0.49\linewidth}
\includegraphics[width=\linewidth]{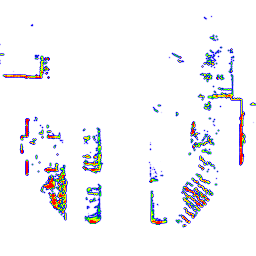}
\caption{Intensity}
\label{fig:intensity}
\end{subfigure}
\begin{subfigure}{0.49\linewidth}
\includegraphics[width=\linewidth]{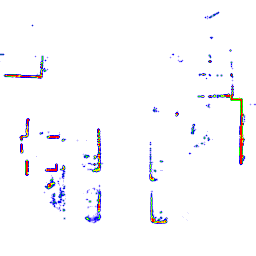}
\caption{Height difference}
\label{fig:height_difference}
\end{subfigure}

\begin{subfigure}{0.49\linewidth}
\includegraphics[width=\linewidth]{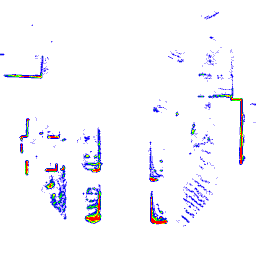}
\caption{Detections}
\label{fig:detections}
\end{subfigure}
\begin{subfigure}{0.49\linewidth}
\includegraphics[width=\linewidth]{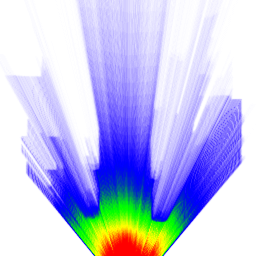}
\caption{Observations}
\label{fig:observations}
\end{subfigure}

\begin{subfigure}{0.49\linewidth}
\includegraphics[width=\linewidth]{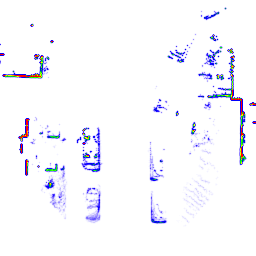}
\caption{Decay rate}
\label{fig:decay_rate}
\end{subfigure}
\begin{subfigure}{0.49\linewidth}
\includegraphics[width=\linewidth]{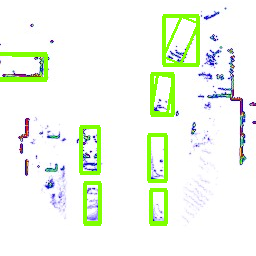}
\caption{Label boxes}
\label{fig:boxes}
\end{subfigure}

\caption{
Grid cell features (Fig.~\ref{fig:intensity}--\ref{fig:decay_rate}) and label boxes (Fig.~\ref{fig:boxes}).
Low values indicated by blue/white color, high values by red color.
The intensity layer carries information on the surface material.
The height difference layer (and consequently the min. / max. z coordinate layer) encodes information that would otherwise be lost due to the projection onto the ground surface.
The number of detections depend on the object distance and their vertical structure while the number of observations describes the observable space.
The decay rate (see \cite{schaefer2017}) is a normalized measure based on detections and observations.
We use rotated and axis-aligned ground truth boxes with semantic classes (Fig.~\ref{fig:boxes}).
Best viewed digitally with zoom.
}
\label{fig:features}
\end{figure}

Exemplary, we follow the approach presented in \cite{schaefer2017} to estimate the decay rate
\begin{equation*}
\lambda_i = \frac{H_i}{\sum_{j \in \mathcal{J}} d_i(j)}
\end{equation*}
for each cell $i$ as the ratio of the number of detections $H_i$ and the sum of distances $d_i(j)$ traveled through $i$ for all rays $j\in\mathcal{J}$.
We determine $\mathcal{J}$ and $d_i$ by casting rays from the sensor origin to end points using the slab method proposed in \cite{kay1986}.
In another configuration, we use the number of detections and observations per cell directly.
To encode height information we use the minimum and maximum z coordinate of all points within a cell instead of splitting the z range into several intervals (e.g. as in \cite{chen2017, ku2017}).
In all configurations we determine the average reflected energy, in the following termed as intensity.
Figure~\ref{fig:features} depicts the grid cell features presented.
Table~\ref{tab:features} summarizes the feature configurations used for evaluation.

\begin{table}[ht]
\centering
\begin{tabular}{c|l}
\textbf{Features Id}	& \textbf{Configuration} 										\\
\hline
\textbf{F1} 			& Intensity, min.\,/\,max. z coord., detections, observations 	\\
\textbf{F2} 			& Intensity, min.\,/\,max. z coord., decay rate					\\
\textbf{F3} 			& Intensity, detections, observations							\\
\textbf{F1*} 		& Same as \textbf{F1} but with ground surface removed				\\
\end{tabular}
\caption{Evaluated feature configurations.}
\label{tab:features}
\end{table}

%% file: sections/04_training.tex
\section{Training}
\label{sec:training}

Out of the total amount of training examples we use 2331 (31\%) samples for internal evaluation, referred to as the validation set.
As summarized in Table~\ref{tab:architectures_results} we train networks with several configurations, varying one parameter at a time.
We pretrain each feature extractor with a learning rate of $1\cdot10^{-3}$ (Resnet) and $6\cdot10^{-4}$ (Inception) for 200k iterations with a grid cell size of 15cm and 10cm, respectively.
A few networks are compared against other methods by uploading inferred labels to the KITTI benchmark.
Due to our limited computational resources we train all networks using SGD, batch normalization \cite{ioffe2015} and use the Momentum optimizer with a momentum of 0.9.
Starting from the trained baseline networks we then train each configuration for another 350k iterations with the initial learning rate lowered by a factor of 2. Then we reduce the learning rate by 10x after 200k and 300k iterations.

\subsection{Box Encoding}
\label{sec:training_box_encoding}
As mentioned in Section~\ref{sec:related_work_box_encoding} there are several box encodings used.
We want to use as few parameters as possible because we assume this to be beneficial for box regression accuracy.
However, while the orientation estimation might be more problematic we adapt the approach in \cite{ku2017} and estimate the orientation $\theta$ by two parameters $\sin(2\theta)$ and $\cos(2\theta)$, providing an explicit and smooth mapping within $\left[-\frac{\pi}{2}, \frac{\pi}{2}\right]$ (B1).
To compare against other encodings we also represent boxes by position, length, width and orientation (B2) as well as two points and width \cite{jiang2017} (B3).
The encodings are summarized in Table~\ref{tab:box_encodings}.

\begin{table}[ht]
\centering
\begin{tabular}{c|l}
\textbf{Box Encoding Id}	& \textbf{Parameters} 												\\
\hline
\textbf{B1} 				& $\left[ x_c, y_c, w, h, \sin(2\theta), \cos(2\theta) \right]$ 		\\
\textbf{B2} 				& $\left[ x_c, y_c, w, h, \theta \right]$ 							\\
\textbf{B3} 				& $\left[ x_1, y_1, x_2, y_2, w \right]$
\end{tabular}
\caption{Evaluated box encodings.}
\label{tab:box_encodings}
\end{table}

\subsection{Data Augmentation}
\label{sec:training_data_augmentation}
Because convolutional filters are not rotationally invariant we increase the amount of training samples by augmenting different viewing angles.
Similar to \cite{chen2017} and \cite{ku2017} we randomly flip the grid map around its x-axis (pointing to the front).
Subsequently, we randomly rotate each grid map within $\left[ \ang{-15}, \ang{15} \right]$ around the sensor origin.
Label boxes are augmented accordingly.

\subsection{Proposal Generation}
\label{sec:training_proposal_generation}
In contrast to \cite{ku2017} we aim to train one network for many classes (see Table~\ref{tab:classes}).
However, as vans and cars as well as sitting persons and pedestrians are similar or only very few training samples are available we merge these classes into one class.

Working on fixed scale grid maps, we can further adapt the object proposal generation to our domain by adapting its size, aspect ratio and stride.
Table~\ref{tab:classes} summarizes the maximum length and width for each semantic class.
Therefore, we determine a small set
\begin{equation*}
\mathcal{S} = \{ 1.75\,m, 2.5\,m, 9\,m, 22\,m \}
\end{equation*}
of anchor sizes that enclose most objects closely.
Note that we determine the combined extent for cars\,/\,vans and pedestrians\,/\,sitting~persons as we treat them to be of the same class.
Trams might not fit completely into the largest feature maps.
However, we think that they can be distinguished properly due to their large size.
We chose the feature slice aspect ratios to be 1:1, 2:1 and 1:2 and the stride to be 16 times the grid cell size.

\subsection{Metrics}
\label{subsec:training_metrics}
To train the RPN we use the same multi-task loss as presented in \cite{ren2017}.
However, for the box classification and regression stage we extend this metric by another sibling output layer and define the multi-task loss similar to \cite{jiang2017} as
\begin{equation*}
L = L_{\mathrm{cls}}(p, p^*) + \lambda_1 L_{\mathrm{loc}, 1}(v, v^*) + \lambda_2 L_{\mathrm{loc}, 2}(u , u^*).
\end{equation*}
For each proposal a discrete probability distribution $p$ over $K+1$ classes is computed by the softmax function.
Here, $L_{\mathrm{cls}}(p, p^*)$ denotes the multi-class cross entropy loss for the true class $p^*$.
$v$ is the predicted bounding-box regression offset given in \cite{ren2017} in which $v$ specifies a scale-invariant translation and log-space height\,/\,width shift relative to an object proposal.
$u$ denotes the predicted inclined bounding-box regression offset.
For the localization losses $L_{\mathrm{loc}, 1}$ and $L_{\mathrm{loc}, 2}$ we use the robust smooth L\textsubscript{1} loss.
Here, $v^*$ denotes the true bounding-box regression target and $u^*$ the true inclined bounding-box regression target depending on the used box encoding (see Table~\ref{tab:box_encodings}).
The hyperparameters $\lambda_1$ and $\lambda_2$ balance the different loss terms and are set to $\lambda_1=\lambda_2=2$ in all experiments.
The difference between the two bounding box representations is also depicted in Figure~\ref{fig:boxes}.

\sisetup{table-number-alignment=center, exponent-product=\cdot, output-decimal-marker = {.}}
\begin{table*}[t]
\centering
\begin{tabular}{c|cc|cc|c|ccc|ccc|ccc|c}
& \multicolumn{2}{c}{\textbf{Architecture}} \vline & \multicolumn{2}{c}{\textbf{Grid Map}} \vline & \multicolumn{1}{c}{\textbf{Box}} \vline & \multicolumn{10}{c}{\textbf{KITTI Evaluation}} \\
\textbf{Net} & \textbf{Meta Arch.} & \textbf{Feat. Extr.} & \textbf{Feat.} & \textbf{Res.} & \textbf{Enc.} & \multicolumn{3}{c}{\textbf{Cars}} \vline & \multicolumn{3}{c}{\textbf{Cyclists}} \vline & \multicolumn{3}{c}{\textbf{Pedestrians}} \vline & \textbf{Time} \\
Id & & & Id & cm & Id & E & M & H & E & M & H & E & M & H & ms\\
\hline
1 & F. R-CNN & Res101 & F1 & 15 & B1 & 66.1 & 50.9 & 50.8 & 11.0 & 10.2 & 10.2 & 19.4 & 16.4 & 16.5 & 93 \\
2 & R-FCN & Res101 & F1 & 15 & B1 & 55.1 & 41.0 & 42.1 & 2.5 & 1.5 & 1.5 & 3.3 & 2.5 & 2.5 & 34 \\        
3 & F. R-CNN & IncV2 & F1 & 15 & B1 & 57.9 & 42.9 & 42.1 & 9.1 & 9.1 & 9.1 & 1.8 & 4.5 & 4.5 & 34 \\
4 & F. R-CNN & Res101 & F2 & 15 & B1 & 70.2 & 53.8 & 53.3 & 13.4 & 9.1 & 9.1 & 32.2 & 26.5 & 26.0 & 92 \\
5 & F. R-CNN & Res101 & F1 & 10 & B1 & 67.3 & 53.3 & 52.8 & 7.7 & 6.5 & 6.1 & 50.4 & 44.0 & 43.5 & 101 \\
6 & F. R-CNN & Res101 & F1 & 15 & B2 & 57.4 & 47.7 & 42.3 & 10.9 & 5.5 & 3.3 & 18.3 & 16.3 & 16.4 & 93 \\
7 & F. R-CNN & Res101 & F3 & 15 & B1 & 70.2 & 54.1 & 52.9 &  9.7 &  9.1 &  9.1 & 2.5 & 2.4 & 2.5 & 91 \\
8 & F. R-CNN & Res101 & F1* & 15 & B1 & 69.8 & 53.3 & 53.1 & 9.7 & 9.1 & 9.1 & 19.3 & 17.6 & 17.5 & 93 \\
\hline
4 & F. R-CNN & Res101 & F2 & 15 & B1 & 70.4 & 55.0 & 47.4 & 14.9 & 11.7 & 11.7 & 15.8 & 12.4 & 12.2 & 92 \\
5 & F. R-CNN & Res101 & F1 & 10 & B1 & 67.5 & 53.7 & 46.5 & 15.7 & 12.5 & 12.8 & 24.3 & 19.1 & 18.5 & 101 \\
\end{tabular}\\
\caption{Object detection configurations with KITTI evaluation results on the validation set (upper part) and on the test set (lower part).
Given the baseline configuration 1, we vary different parameters, one at a time.
}
\label{tab:architectures_results}
\end{table*}

%% file: sections/05_evaluation.tex
\section{Evaluation}
\label{sec:evaluation}

Table~\ref{tab:architectures_results} summarizes the evaluation results on the validation and test set for different network configurations.

\subsection{Metrics}
\label{sec:evaluation_metrics}
We evaluate the overall accuracy based on the average precision for the KITTI Bird's Eye View Evaluation using an Intersection over Union (IoU) threshold of 0.7 for cars and an IoU of 0.5 for cyclists and pedestrians.
The evaluation is divided into the three difficulties Easy\,(E), Moderate\,(M) and Hard\,(H) based on occlusion level, maximal truncation and minimum bounding box size.

\subsection{Accuracy}
\label{sec:evaluation_accuracy}
Table~\ref{tab:architectures_results} summarizes the quantitative evaluation results using the KITTI benchmark metric.

\begin{figure}
\begin{scaletikzpicturetowidth}{\columnwidth}
\begin{tikzpicture}[scale=\tikzscale]
\begin{axis}[
	enlargelimits=true,
	xlabel=IoU / \%,
	ylabel=Accuracy / \%,
	ytick={0,10,...,90},
	grid=both,
	x label style={at={(axis description cs:.5 ,-.05)}, anchor=south, font=\small},
	y label style={at={(axis description cs:0.075 ,0.3)}, anchor=west, font=\small},
    tick label style={font=\small},
    cycle list name=exotic,
    legend style={anchor=north, font=\small},
]

\addplot plot coordinates {(20, 87.8)(30, 87.8)(40, 87.8)(50, 87.0)(60, 84.8)(70, 67.3)(80, 31.6)};
\addplot plot coordinates {(20, 32.4)(30, 21.0)(40, 13.3)(50, 7.7)(60, 4.5)(70, 4.5)(80, 0)};
\addplot plot coordinates {(20, 75.9)(30, 73.8)(40, 66.0)(50, 50.4)(60, 22.1)(70, 3.7)(80, 0.5)};

\legend{Cars\\Cyclists\\Pedestrians\\}
\end{axis}
\end{tikzpicture}
\end{scaletikzpicturetowidth}

\caption{Overall validation accuracy depending on the Intersection-over-Union (IoU) for Net 5 in the easy benchmark.}
\label{fig:accuracy_vs_iou}
\end{figure}
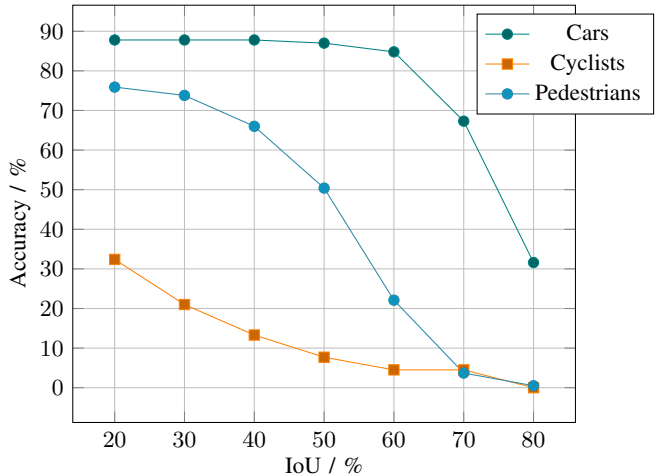

The largest gain in accuracy for smaller objects such as cyclists and pedestrians is made by decreasing the grid cell size as for Net\,5.
However, also the box encoding has a large impact on the accuracy.
While in Net\,6 angles can not be recovered robustly the angle encoding in B\,1 yields better results.
Unfortunately, the network training for box encoding B3 did not converge at all.
This might be due to an issue during data augmentation when boxes (and grid maps) are rotated.
Also, the input features have an impact on the detection accuracy.
It seems that normalization via the decay rate model yields better results than using the number of detections and observations directly.
This is advantageous as the amount of grid map layers can be decreased this way.
Ground surface removal has a minor impact on the detection of cars and other large objects but leads to a reduced accuracy in the detection of cyclists and pedestrians.
We believe that this is due to detections close to the ground surface that are removed.

Figure~\ref{fig:accuracy_vs_iou} depicts the accuracy depending on the Intersection-over-Union (IoU) for Net 5 in the easy benchmark where it achieves a comparably high accuracy up to a desired IoU of about 50\%.

Our test results (submitted as \textit{TopNet} variants) are mostly similar to the validation results, yielding state-of-the-art benchmark results, likely due to our data augmentation strategies.
For pedestrians the accuracy on the test set dropped.
The reason might be overfitting due to a non-optimal validation data split for pedestrians.

\begin{figure*}
\centering
\begin{subfigure}[h]{0.32\linewidth}
\includegraphics[width=\linewidth]{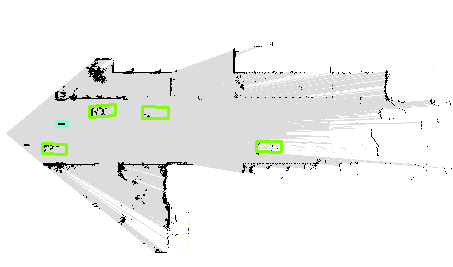}
\end{subfigure}
\begin{subfigure}[h]{0.32\linewidth}
\includegraphics[width=\linewidth]{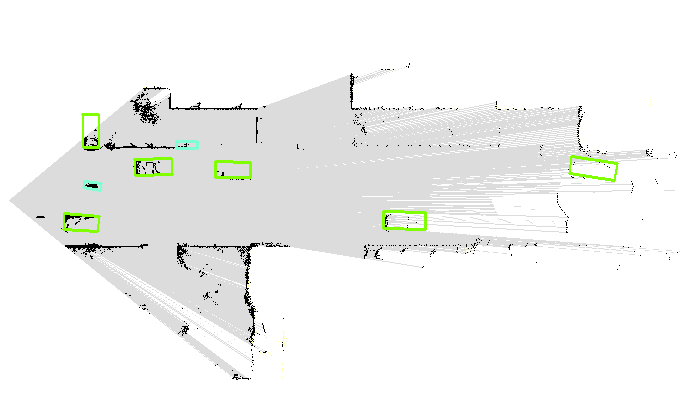}
\end{subfigure}
\begin{subfigure}[h]{0.32\linewidth}
\includegraphics[width=\linewidth]{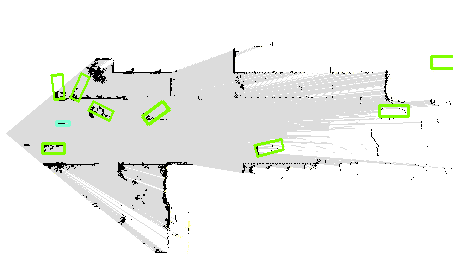}
\end{subfigure}

\begin{subfigure}[h]{0.32\linewidth}
\includegraphics[width=\linewidth]{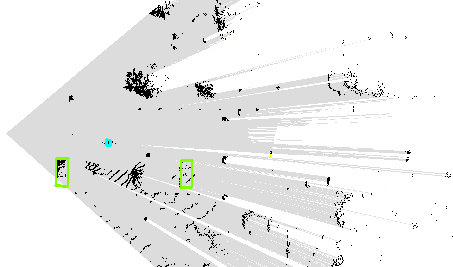}
\caption{Net\,1}
\end{subfigure}
\begin{subfigure}[h]{0.32\linewidth}
\includegraphics[width=\linewidth]{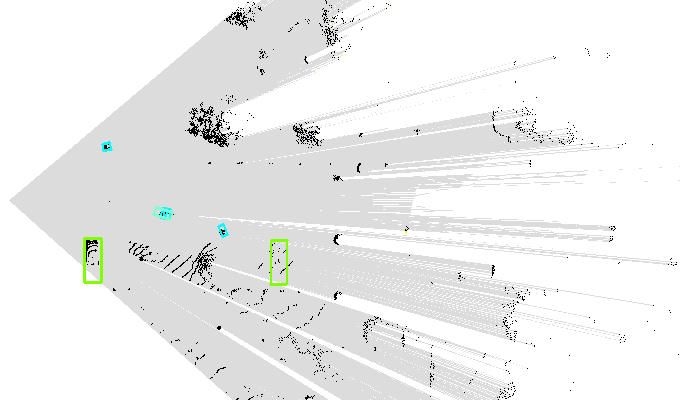}
\caption{Net\,5}
\end{subfigure}
\begin{subfigure}[h]{0.32\linewidth}
\includegraphics[width=\linewidth]{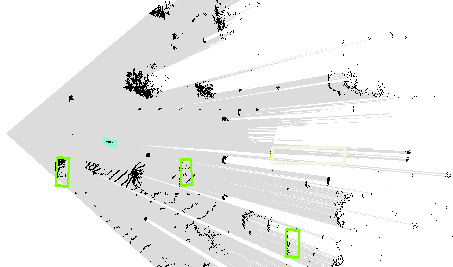}
\caption{Net\,6}
\end{subfigure}

\caption{
Qualitative results for three different networks on two scenarios.
Cars are depicted green, cyclists aquamarin and pedestrians cyan.
Compared to Net\,1, Net\,5 is able to detect pedestrians and distant cars.
Due to the box encoding in Net\,6 the rotation regression is less robust compared to Net\,1 and Net\,5.
Best viewed digitally with zoom.
}
\label{fig:qualitative_results}
\end{figure*}

Figure~\ref{fig:qualitative_results} depicts two scenarios for qualitative comparison of three network configurations.

\subsection{Inference Time}
\label{sec:evaluation_inference_time}
We evaluated the processing times on a 2.5 GHz six core Intel Xeon E5-2640 CPU with 15 MB cache and an NVIDIA GeForce GTX 1080 Ti GPU with 11 GB graphics memory.
In comparison to the other networks Net\,5 has the highest inference time.
Net\,2 has a considerably shorter inference time due to the R-FCN meta architecture.
Using the InceptionV2 feature extractor in Net\,3 also drastically decreases the inference time compared to using a Resnet101.
A different number of grid map layers or different box encodings have no significant impact on the inference time.

%% file: sections/06_conclusion.tex
\section{Conclusion}
\label{sec:conclusion}

We presented our approach to object detection and classification based on multi-layer grid maps using deep convolutional networks.

By specifically adapting preprocessing, input features, data augmentation, object encodings and proposal generation to the grid map domain we show that our networks achieve state of the art benchmark results by only using multi-layer grid maps from range sensor data.
We identify the input feature selection together with the resolution as an important factor for network accuracy and training\,/\,inference time.

As a next step we aim to develop a framework for semi-supervised learning of object detectors, hopefully increasing generalization and thus overall robustness.
Finally, we want to develop a tracking framework based on grid maps by coupling detections with predictions in an end-to-end learnable framework.

%% file: root.bbl
\begin{thebibliography}{10}
\providecommand{\url}[1]{#1}
\csname url@samestyle\endcsname
\providecommand{\newblock}{\relax}
\providecommand{\bibinfo}[2]{#2}
\providecommand{\BIBentrySTDinterwordspacing}{\spaceskip=0pt\relax}
\providecommand{\BIBentryALTinterwordstretchfactor}{4}
\providecommand{\BIBentryALTinterwordspacing}{\spaceskip=\fontdimen2\font plus
\BIBentryALTinterwordstretchfactor\fontdimen3\font minus
  \fontdimen4\font\relax}
\providecommand{\BIBforeignlanguage}[2]{{%
\expandafter\ifx\csname l@#1\endcsname\relax
\typeout{** WARNING: IEEEtran.bst: No hyphenation pattern has been}%
\typeout{** loaded for the language `#1'. Using the pattern for}%
\typeout{** the default language instead.}%
\else
\language=\csname l@#1\endcsname
\fi
#2}}
\providecommand{\BIBdecl}{\relax}
\BIBdecl

\bibitem{elfes1989}
A.~Elfes, ``{{Using Occupancy Grids for Mobile Robot Perception and
  Navigation}},'' \emph{Computer}, vol.~22, no.~6, pp. 46--57, 1989.

\bibitem{nuss2015}
D.~Nuss, T.~Yuan, G.~Krehl, M.~Stuebler, S.~Reuter, and K.~Dietmayer,
  ``{{Fusion of Laser and Radar Sensor Data with a Sequential Monte Carlo
  Bayesian Occupancy Filter}},'' in \emph{IEEE Intelligent Vehicles Symposium
  (IV)}, 2015, pp. 1074--1081.

\bibitem{ren2017}
S.~Ren, K.~He, R.~Girshick, and J.~Sun, ``{{Faster R-CNN: Towards Real-Time
  Object Detection with Region Proposal Networks}},'' \emph{IEEE Transactions
  on Pattern Analysis and Machine Intelligence}, vol.~39, no.~6, pp.
  1137--1149, 2017.

\bibitem{dai2016}
J.~Dai, Y.~Li, K.~He, and J.~Sun, ``{{R-FCN: Object Detection via Region-based
  Fully Convolutional Networks}},'' 2016.

\bibitem{liu2015}
W.~Liu, D.~Anguelov, D.~Erhan, C.~Szegedy, S.~Reed, C.-Y. Fu, and A.~C. Berg,
  ``{{SSD: Single Shot MultiBox Detector}},'' 2015.

\bibitem{he2016}
K.~He, X.~Zhang, S.~Ren, and J.~Sun, ``{{Deep Residual Learning for Image
  Recognition}},'' in \emph{IEEE Conference on Computer Vision and Pattern
  Recognition}, 2016, pp. 770--778.

\bibitem{szegedy2016}
C.~Szegedy, S.~Ioffe, V.~Vanhoucke, and A.~Alemi, ``{{Inception-v4,
  Inception-ResNet and the Impact of Residual Connections on Learning}},''
  2016.

\bibitem{howard2017}
A.~G. Howard, M.~Zhu, B.~Chen, D.~Kalenichenko, W.~Wang, T.~Weyand,
  M.~Andreetto, and H.~Adam, ``{{MobileNets: Efficient Convolutional Neural
  Networks for Mobile Vision Applications}},'' 2017.

\bibitem{hinz2003b}
S.~Hinz, ``{{Detection and Counting of Cars in Aerial Images}},'' in
  \emph{International Conference on Image Processing}, vol.~3, 2003, pp.
  III--997--1000 vol.2.

\bibitem{nguyen2007}
T.~T. Nguyen, H.~Grabner, H.~Bischof, and B.~Gruber, ``{{On-line Boosting for
  Car Detection from Aerial Images}},'' in \emph{IEEE International Conference
  on Research, Innovation and Vision for the Future}, 2007, pp. 87--95.

\bibitem{kluckner2007}
S.~Kluckner, G.~Pacher, H.~Grabner, H.~Bischof, and J.~Bauer, ``{{A 3D Teacher
  for Car Detection in Aerial Images}},'' in \emph{2007 IEEE 11th International
  Conference on Computer Vision}, 2007, pp. 1--8.

\bibitem{wirges2018}
S.~Wirges, F.~Hartenbach, and C.~Stiller, ``{{Evidential Occupancy Grid Map
  Augmentation using Deep Learning}},'' 2018.

\bibitem{zhao2003}
T.~Zhao and R.~Nevatia, ``{{Car Detection in Low Resolution Aerial Images}},''
  \emph{Image and Vision Computing}, vol.~21, no.~8, pp. 693--703, 2003.

\bibitem{geiger2012}
A.~Geiger, P.~Lenz, and R.~Urtasun, ``{{Are we Ready for Autonomous Driving?
  The KITTI Vision Benchmark Suite}},'' \emph{IEEE Conference on Computer
  Vision and Pattern Recognition}, pp. 3354--3361, 2012.

\bibitem{chen2017}
X.~Chen, H.~Ma, J.~Wan, B.~Li, and T.~Xia, ``{{Multi-View 3D Object Detection
  Network for Autonomous Driving}},'' \emph{CVPR}, pp. 1907--1915, 2017.

\bibitem{ku2017}
J.~Ku, M.~Mozifian, J.~Lee, A.~Harakeh, and S.~Waslander, ``{{Joint 3D Proposal
  Generation and Object Detection from View Aggregation}},'' 2017.

\bibitem{golovinskiy2009}
A.~Golovinskiy, V.~G. Kim, and T.~Funkhouser, ``{{Shape-based Recognition of 3d
  Point Clouds in Urban Environments}},'' \emph{IEEE International Conference
  on Computer Vision}, pp. 2154--2161, 2009.

\bibitem{babahajiani2015}
P.~Babahajiani, L.~Fan, and M.~Gabbouj, ``{{Object Recognition in 3D Point
  Cloud of Urban Street Scene}},'' C.~V. Jawahar and S.~Shan, Eds.\hskip 1em
  plus 0.5em minus 0.4em\relax Cham: Springer International Publishing, 2015,
  pp. 177--190.

\bibitem{hoermann2018}
S.~Hoermann, P.~Henzler, M.~Bach, and K.~Dietmayer, ``{{Object Detection on
  Dynamic Occupancy Grid Maps Using Deep Learning and Automatic Label
  Generation}},'' 2018.

\bibitem{jiang2017}
Y.~Jiang, X.~Zhu, X.~Wang, S.~Yang, W.~Li, H.~Wang, P.~Fu, and Z.~Luo,
  ``{{R2CNN: Rotational Region CNN for Orientation Robust Scene Text
  Detection}},'' 2017.

\bibitem{ceres}
S.~Agarwal, K.~Mierle, and Others, ``{{Ceres Solver}},''
  \url{http://ceres-solver.org}.

\bibitem{schaefer2017}
A.~Schaefer, L.~Luft, and W.~Burgard, ``{{An Analytical Lidar Sensor Model
  Based on Ray Path Information}},'' \emph{IEEE International Conference on
  Robotics and Automation}, vol.~2, no.~3, pp. 1405--1412, 2017.

\bibitem{kay1986}
T.~L. Kay and J.~T. Kajiya, ``{{{Ray Tracing Complex Scenes}}},'' in \emph{ACM
  SIGGRAPH Computer Graphics}, vol.~20, no.~4, 1986, pp. 269--278.

\bibitem{ioffe2015}
S.~Ioffe and C.~Szegedy, ``{{Batch Normalization: Accelerating Deep Network
  Training by Reducing Internal Covariate Shift}},'' 2015.

\end{thebibliography}
